# Compositional Semantics Grounded in Commonsense Metaphysics

Walid S. Saba

American Institutes for Research, 1000 Thomas Jefferson St., Washington, DC 20007 USA
wsaba@air.org

**Abstract.** We argue for a compositional semantics grounded in a strongly typed ontology that reflects our commonsense view of the world and the way we talk about it in ordinary language. Assuming the existence of such a structure, we show that the semantics of various natural language phenomena may become nearly trivial.

## 1 Introduction

If the main business of semantics is to explain how linguistic constructs relate to the world, then semantic analysis of natural language text is, indirectly, an attempt at uncovering the semiotic ontology of commonsense knowledge, and particularly the background knowledge that seems to be implicit in all that we say in our everyday discourse. While this intimate relationship between language and the world is generally accepted, semantics (in all its paradigms) has traditionally proceeded in one direction: by first stipulating an assumed set of ontological commitments followed by some machinery that is supposed to, somehow, model meanings in terms of that stipulated structure of reality.

With the gross mismatch between the trivial ontological commitments of our semantic formalisms and the reality of the world these formalisms purport to represent, it is not surprising therefore that challenges in the semantics of natural language are rampant. However, as correctly observed in [7], semantics could become nearly trivial if it was grounded in an ontological structure that is, "isomorphic to the way we talk about the world". The obvious question however is 'how does one arrive at this ontological structure that implicitly underlies all that we say in everyday discourse?' One plausible answer is the (seemingly circular) suggestion that the semantic analysis of natural language should itself be used to uncover this structure. In this regard we strongly agree with [4] who states:

> We must not try to resolve the metaphysical questions first, and then construct a meaning-theory in light of the answers. We should investigate how our language actually functions, and how we can construct a workable systematic description of how it functions; the answers to those questions will then determine the answers to the metaphysical ones.

What this suggests, and correctly so, in our opinion, is that in our effort to understand the complex and intimate relationship between ordinary language and everyday commonsense knowledge, one could, as also suggested in [2], "use language as a tool for uncovering the semiotic ontology of commonsense" since ordinary language is the best known theory we have of everyday knowledge.

To avoid this seeming circularity (in wanting this ontological structure that would trivialize semantics; while at the same time suggesting that semantic analysis should itself be used as a guide to uncovering this ontological structure), we could start performing semantic analysis from the ground up, assuming a minimal (almost a trivial and basic) ontology, building up the ontology as we go guided by the results of the semantic analysis. The advantages of this approach are: (*i*) the ontology thus constructed as a result of this process would not be invented, as is the case in most approaches to ontology (e.g., [5], [8] and [13]), but would instead be discovered from what is in fact implicitly assumed in our use of language in everyday discourse; (*ii*) the semantics of several natural language phenomena should as a result become trivial, since the semantic analysis was itself the source of the underlying knowledge structures (in a sense, the semantics would have been done before we even started!)

In this paper we suggest exactly such an approach. In particular, in the rest of the paper we (*i*) argue that semantics must be grounded in a much richer ontological structure, one that reflects our commonsense view of the world and the way we talk about it in ordinary language; (*ii*) it will be demonstrated that in a logic 'embedded' with commonsense metaphysics the semantics of various natural language phenomena could become 'nearly' trivial; and (*iii*) we finally suggest some steps towards discovering (as opposed to inventing) the ontological structure that seems to implicitly underlie all that we say in ordinary language.

## 2 Semantics with Ontological Content

We begin by making a case for a semantics that is grounded in a strongly typed ontological structure that is isomorphic to our commonsense view of reality. In doing so, our ontological commitments will initially be minimal. In particular, we assume the existence of a subsumption hierarchy of a number of general categories such as `animal`, `substance`, `entity`, `artifact`, `event`, etc., and where the fact that an object of type `human` is also an `entity`, for example, is expressed as $\text{human} \sqsubseteq \text{entity}$. We shall use $(x :: \text{animal})$ to state that $x$ is an object of type `animal`, and $Articulate(x :: \text{human})$ to state that the property *Articulate* is true of some object $x$, an object that must be of type `human` (since 'articulate' is a property that is ordinarily said of humans). We write $(\exists x :: \text{t})(P(x))$ when the property $P$ is true of some object $x$ of type $\text{t}$; $(\exists^1 x :: \text{t})(P(x))$ when $P$ is true of a unique object of type $\text{t}$; and $(\exists x :: \text{t}^a)(P(x))$ when the property $P$ is true of some object $x$ of type $\text{t}$, an object that only conceptually (or abstractly) exists - i.e., an object that need not physically exist. Furthermore, we assume $(Qx :: \text{t})(P(x)) \supset (Qx :: \text{t}^a)(P(x))$, where $Q$ is one of the standard quantifiers $\forall$ and $\exists$, i.e., what actually exists must also conceptually exist. Proper nouns, such as *Sheba*, are interpreted as

$$\|sheba\| \Rightarrow \lambda P[(\exists^1 x)(Noo(x::\texttt{entity},\text{'}sheba\text{'}) \wedge P(x::\texttt{t}))] \tag{1}$$

where $Noo(x::\texttt{entity}, s)$ is true of some individual object $x$ (which could be any $\texttt{entity}$), and $s$ if (the label) $s$ is the name of $x$ (to simplify notation we sometimes write (1) as $\|sheba\| \Rightarrow \lambda P[(\exists^1 x :: sheba)(P(x::\texttt{t}))]$). Note now that a variable might, in a single scope, be associated with more than one type. For example, $x$ in (1) is considered to be an $\texttt{entity}$ and an object of type $\texttt{t}$, where $\texttt{t}$ is presumably the type of objects that the property $P$ applies to (or makes sense of). In these situations a type unification must occur. In particular, type unification occurs when some variable $x$ is associated with more than one type in a single scope. A type unification $(\texttt{s} \bullet \texttt{t})$, between two types $\texttt{s}$ and $\texttt{t}$, where $Q$ is one of the standard quantifiers $\forall$ and $\exists$ is defined as follows:

$$(Qx :: (\texttt{s} \bullet \texttt{t}))(P(x)) \equiv \begin{cases} (Qx :: \texttt{s})(P(x)), & \text{if } (\texttt{s} \sqsubseteq \texttt{t}) \\ (Qx :: \texttt{t})(P(x)), & \text{if } (\texttt{t} \sqsubseteq \texttt{s}) \\ (Qx :: \texttt{s})(\exists y :: \texttt{t}^a)(R(x,y) \wedge P(y)), & \text{if } (\exists R)(R(x,y)) \\ (Qx :: \bot)(P(x)), & \text{otherwise} \end{cases} \tag{2}$$

As an initial example, consider the steps involved in the interpretation of '*sheba is hungry*', where it will be assumed that ($\texttt{animal} \sqsubseteq \texttt{entity}$) and that *Hungry* is a property that applies to (or makes sense of) objects that are of type $\texttt{animal}$:

$\|sheba\text{ is hungry}\|$ (3)

$\Rightarrow (\exists^1 sheba :: \texttt{entity})(Hungry(sheba :: \texttt{animal}))$

$\Rightarrow (\exists^1 sheba :: (\texttt{animal} \bullet \texttt{entity}))(Hungry(sheba))$

$\Rightarrow (\exists^1 sheba :: \texttt{animal})(Hungry(sheba))$

Thus, '*sheba is hungry*' states that there is a unique object named *sheba*, which must be an object of type $\texttt{animal}$, and such that *sheba* is *hungry*. Type unification will not always be as straightforward, as in general there could be more than two types associated with a variable in a single scope. Consider for example the interpretation of '*sheba is a young artist*', where *Young* is assumed to be a property that applies to (or makes sense of) $\texttt{physical}$ objects; and where it is assumed that *Artist* is a property that is ordinarily said of objects that are of type $\texttt{human}$:

$\|sheba\text{ is a young artist}\|$ (4)

$\Rightarrow (\exists^1 sheba :: \texttt{entity})(Artist(sheba :: \texttt{human}) \wedge Young(sheba :: \texttt{physical}))$

A pair of type unifications, must now occur: $(\texttt{entity} \bullet (\texttt{human} \bullet \texttt{physical}))$, resulting in $\texttt{human}$. The final interpretation is thus given as follows:

$\|sheba\text{ is a young artist}\|$ (5)

$\Rightarrow (\exists^1 sheba :: \texttt{human})(Artist(sheba) \wedge Young(sheba))$

In the final analysis, therefore, '*sheba is a young artist*' is interpreted as follows: there is a unique object named *sheba*, an object that must be of type `human`, and such that *sheba* is *Artist* and *Young*. It should be noted here that not recognizing the ontological difference between `human` and *Artist* (namely, that what ontologically exist are objects of type `human`, and not artists, and that *Artist* is a mere property that may or may not apply to objects of type `human`) has traditionally led to ontologies rampant with multiple inheritance. Note, further, that in contrast with `human`, which is a first-intension ontological concept (see [3] for a formal discussion on this issue), *Artist* and *Young* are considered to be second-intension logical concepts, namely properties that may or may not be true of first-intension (ontological) concepts. Moreover, and unlike first-intension ontological concepts (such as `human`), logical concepts such as *Artist* are assumed to be defined by virtue of logical expressions, such as $(\forall x :: \texttt{human})(Artist(x) \equiv_{df} \varphi)$, where the exact nature of $\varphi$ might very well be susceptible to temporal, cultural, and other contextual factors, depending on what, at a certain point in time, a certain community considers an *Artist* to be. That is, while the properties of being an *Artist* and *Young* that *x* exhibits are accidental (as well as temporal, cultural-dependent, etc.), the fact that some *x* is `human` (and thus an `animal`, etc.) is not[1].

## 3  More on Type Unification

Thus far we performed simple type unifications involving types that are in a subsumption relationship. For example, we assumed $(\texttt{human} \bullet \texttt{entity}) = \texttt{human}$, since $(\texttt{human} \sqsubseteq \texttt{entity})$. Quite often, however, it is not subsumption but some other relationship that exists between the different types associated with a variable, and a typical example is the case of nominal compounds. Consider the following:

a. *book review* (6)
b. *book proposal*
c. *design review*
d. *design plan*

From the standpoint of commonsense, the reference to a `book review` should imply the existence of a `book`, whereas the reference to a `book proposal` should be considered to be a reference to a `proposal` of some `book`, a book that might not (yet) actually exist. That is,

---

[1] In a recent argument *Against Fantology* [12] it was noted that too much attention has been paid to the false doctrine that much can be discovered about the ontological structure of reality by predication in first-order logic. In particular, it is argued in [12] that the use of standard predication in first-order logic in representing the meanings of '*John is a human*' and '*John is tall*', for example, completely ignores the different ontological categories implicit in each utterance. While we agree with this observation, we believe that our approach to a semantics grounded in an a rich ontological structure that is supposed to reflect our commonsense reality, does solve this problem without introducing ad-hoc relations to the formalism, as example (4) and subsequent examples in this paper demonstrate. First-order logic (and Frege, for that matter), are therefore not necessarily the villains, and the "predicates do not represent" slogan is perhaps appropriate, but it seems only when predicates are devoid of any ontological content.

$$\|a\ book\ review\| \Rightarrow \lambda P[(\exists x :: \texttt{book})(\exists y :: \texttt{review})(ReviewOf(y,x) \wedge P(y))] \quad (7)$$

$$\|a\ book\ proposal\| \Rightarrow \lambda P[(\exists x :: \texttt{book})(\exists y :: \texttt{proposal})(ProposalFor(y,x) \wedge P(y))] \quad (8)$$

Finally, it must be noted that, in general, type unification might fail, and this occurs in the absence of any relationship between the types assigned to a variable in the same scope. For example, assuming $Artificial(x :: \texttt{naturalObj})$, i.e., that $Artificial$ is a property ordinarily said of objects of type $\texttt{naturalObj}$, and assuming $(\texttt{car} \sqsubseteq \texttt{artifact})$, then '*artificial car*' would get the interpretation

$$\|an\ artificial\ car\| \quad (9)$$
$$\Rightarrow \lambda P[(\exists x :: \texttt{car})(Artificial(x :: \texttt{naturalObj}))]$$
$$\Rightarrow \lambda P[(\exists x :: (\texttt{naturalObj} \bullet \texttt{car}))(Artificial(x))]$$
$$\Rightarrow \lambda P[(\exists x :: \bot)(Artificial(x))]$$

It would seem therefore that type unification fails in the interpretation of some phrase that does not seem to be plausible from the standpoint of commonsense[2].

## 4  From Abstract to Actual Existence

What we have been doing thus far can be summarized as follows: we have embedded commonsense into our semantics by annotating every quantified variable referred to in some predicate $P$ with an ontological category that $P$ applies to (or makes sense of), as per our everyday use of ordinary language. In this section it will be demonstrated how this mechanism, along with the notion of type unification, can explain how certain abstract entities that initially can only be assumed to conceptually exist, are, in an appropriate context, reduced to concrete spatio-temporal entities.

Recall that our intention in associating types with quantified variables, as, for example, in $Articulate(x :: \texttt{human})$, was to reflect our commonsense understanding of how the property $Articulate$ is used in our everyday discourse, namely that $Articulate$ is ordinarily said of objects that are of type human. What of a property such as Imminent, then? Undoubtedly, saying some object $e$ is Imminent only makes sense in ordinary language when $e$ is some event, which we have been expressing as $Imminent(e :: \texttt{event})$. But there is obviously more that we can assume of $e$. In particular, imminent is said in ordinary language of some $e$ when $e$ is an event that has

---

[2]  Interestingly, type unification and the embedding of ontological types into our semantics seems also promising in providing an explanation for the notion of metonymy in natural language. While we cannot get into this issue here in much details, we will simply consider the following example by way of illustration, where $R$ is some salient relationship between a $\texttt{human}$ and a $\texttt{hamSandwich}$:

$\|the\ ham\ sandwich\ ordered\ a\ beer\|$
$\Rightarrow (\exists^1 x :: \texttt{hamSandwich})(\exists y :: \texttt{beer})(Ordered(x :: \texttt{human}, y :: \texttt{order}))$
$\Rightarrow (\exists^1 x :: (\texttt{hamSandwich} \bullet \texttt{human}))(\exists y :: (\texttt{beer} \bullet \texttt{object}))(Ordered(x,y))$
$\Rightarrow (\exists^1 z :: \texttt{human}))(\exists^1 x :: \texttt{hamSandwich}^a)(\exists y :: \texttt{beer})(R(x,y) \wedge Ordered(x,y))$

That is, saying 'the ham sandwich ordered a beer' essentially means (again, as far as commonsense is concerned) that some $\texttt{human}$, who stands in some relationship to a $\texttt{hamSandwich}$, ordered a beer.

not yet occurred, that is, an event that exists only conceptually, which we write as $Imminent(e::\texttt{event}^a)$. A question that arises now is this: what is the status of an event *e* that, at the same time, is imminent as well as important? Clearly, an important and imminent event should still be assumed to be an event that does not actually exist (as important as it may be). '*Important*' must therefore be a property that is said of an event that also need not actually exist, as illustrated by the following:

(10)
$\|\textit{an important and imminent event}\|$

$\Rightarrow \lambda P[(\exists x)(Importnat(x::\texttt{abstract}^a) \wedge Imminent(x::\texttt{event}^a) \wedge P(x::\texttt{t}))]$

$\Rightarrow \lambda P[(\exists x::(\texttt{event}^a \bullet \texttt{abstract}^a))(Importnat(x) \wedge Imminent(x) \wedge P(x::\texttt{t}))]$

$\Rightarrow \lambda P[(\exists x::\texttt{event}^a)(Importnat(x) \wedge Imminent(x) \wedge P(x::\texttt{t}))]$

It is important to note here that one can always 'bring down' an object (such as an `event`) from abstract existence into actual existence, but the reverse is not true. Consequently, quantification over variables associated with the type of an abstract concept, such as `event`, should always initially assume abstract existence. To illustrate, let us first assume the following:

(11)
$Attend(x::\texttt{human}, y::\texttt{event})$

(12)
$Cancel(x::\texttt{human}, y::\texttt{event}^a)$

That is, we have assumed that it always makes sense to speak of a `human` that attended or cancelled some `event`, where to attend an event is to have an existing event; and where the object of a cancellation is an `event` that does not (anymore, if it ever did) exist[3]. Consider now the following:

(13)
$\|\textit{john attended the seminar}\|$

$\Rightarrow (\exists^1 j::\texttt{human})(\exists^1 e::\texttt{seminar}^a)(Attended(j::\texttt{human}, e::\texttt{event}))$

$\Rightarrow (\exists^1 j::(\texttt{human} \bullet \texttt{human}))(\exists^1 e::(\texttt{seminar}^a \bullet \texttt{event}))(Attended(j,e))$

$\Rightarrow (\exists^1 j::\texttt{human})(\exists^1 e::\texttt{seminar})(Attended(j,e))$

That is, saying '*john attended the seminar*' is saying there is a specific `human` named *j*, a specific `seminar` *e* (that actually exists) such that *j* attended *e*. On the other hand, consider now the interpretation of the sentence in (14).

(14)
$\|\textit{john cancelled the seminar}\|$

$\Rightarrow (\exists^1 john::\texttt{human})(\exists^1 y::\texttt{seminar}^a)(Cancelled(john::\texttt{human}, y::\texttt{event}^a))$

$\Rightarrow (\exists^1 john::(\texttt{human} \bullet \texttt{human}))(\exists^1 y::(\texttt{seminar}^a \bullet \texttt{event}^a))(Cancelled(john,y))$

$\Rightarrow (\exists^1 john::\texttt{human})(\exists^1 y::\texttt{seminar}^a)(Cancelled(john,y))$

---

[3] Tense and modal aspects can also effect the initial type assignments, although a full treatment of this issue would involve discussing the interaction with syntax in much more detail.

What (14) states is that there is a specific human named *john*, and a specific seminar (that does not necessarily exist), a seminar that *john* cancelled[4]. An interesting case now occurs when a type is 'brought down' from abstract existence into actual existence. Let us assume $Plan(x::\text{human}, y::\text{event}^a)$; that is, that it always makes sense to say that some human is planning (or did plan) an event that need not (yet) actually exist. Consider now the following,

$$\|john\ planned\ the\ trip\| \tag{15}$$

$\Rightarrow (\exists^1 j::\text{human})(\exists^1 e::\text{trip}^a)(Planned(x::\text{human}, y::\text{event}^a))$

$\Rightarrow (\exists^1 j::(\text{human} \bullet \text{human}))(\exists^1 e::(\text{trip}^a \bullet \text{event}^a))(Planned(j,e))$

$\Rightarrow (\exists^1 j::\text{human})(\exists^1 e::\text{trip}^a)(Planned(j,e))$

That is, saying '*john planned the trip*' is simply saying that a specific object that must be a human has planned a specific trip, a trip that might not have actually happened[5]. However, assuming $Lengthy(e::\text{event})$; i.e., that *Lengthy* is a property that is ordinarily said of an (existing) event, then the interpretation of '*john planned the lengthy trip*' should proceed as follows:

$$\|john\ planned\ the\ lengthy\ trip\| \tag{16}$$

$\Rightarrow (\exists^1 j::\text{human})(\exists^1 e::\text{trip}^a)(Planned(x::\text{human}, y::\text{event}^a) \wedge Lengthy(e::\text{event}))$

$\Rightarrow (\exists^1 j::\text{human})(\exists^1 e::\text{trip}^a)(Planned(j,e::(\text{event} \bullet \text{event}^a)) \wedge Lengthy(e))$

$\Rightarrow (\exists^1 j::\text{human})(\exists^1 e::(\text{trip}^a \bullet \text{event}))(Planned(j,e) \wedge Lengthy(e))$

$\Rightarrow (\exists^1 j::\text{human})(\exists^1 e::\text{trip})(Planned(j,e) \wedge Lengthy(e))$

That is, there is a specific human (named *john*) that has planned a specific trip, a trip that was *Lengthy*. It should be noted here that the trip in (16) was finally considered to be an existing event due to other information contained in the same sentence. In general, however, this information can be contained in a larger discourse. For example, in interpreting '*John planned the trip. It was lengthy*' the resolution of 'it' would force a retraction of the types inferred in processing '*John planned the trip*', as the information that follows will 'bring down' the aforementioned trip from abstract to actual existence. This subject is clearly beyond the scope of this paper, but readers interested in the computational details of such processes are referred to [14].

## 5 On an Intensional Verbs and Dot (•) Objects

Consider the following sentences and their corresponding translation into standard first-order logic:

---

[4] As correctly noted in [6], assuming that the reference to the seminar is intensional, i.e., that the reference is to 'the idea of a seminar' does not solve the problem since it is not the idea of a seminar that was cancelled, but an actual event that did not actually happen!

[5] Note that it is the trip (event) that did not necessarily happen, not the planning (activity) for it.

$$\|john\ found\ a\ unicorn\| \Rightarrow (\exists x)(Unicorn(x) \wedge Found(j,x)) \tag{18}$$

$$\|john\ sought\ a\ unicorn\| \Rightarrow (\exists x)(Unicorn(x) \wedge Sought(j,x)) \tag{19}$$

Note that $(\exists x)(Elephant(x))$ can be inferred in both cases, although it is clear that '*john sought a unicorn*' should not entail the existence of a unicorn. In addressing this problem, [9] suggested a solution that in effect treats 'seek' as an intensional verb that has more or less the meaning of `tries to find', using the tools of a higher-order intensional logic. In addition to unnecessary complication of the logical form, however, we believe that this is, at best, a partial solution since the problem in our opinion is not necessarily in the verb *seek*, nor in the reference to unicorns. That is, *painting*, *imagining*, etc. of a unicorn (or an elephant, for that matter) should not entail the existence of a unicorn (nor the existence of an elephant). To illustrate further, let us first assume the following:

$$Paint(x::\texttt{human}, y::\texttt{painting}) \tag{20}$$
$$Find(x::\texttt{human}, y::\texttt{entity}) \tag{21}$$

That is, we are assuming that it always makes sense to speak of a `human` that painted some `painting`, and of some `human` that found some `entity`. Consider now the interpretation in (22), where it was assumed that *Large* is a property that applies to (or makes sense of) objects that are of type `physical`.

$$\|john\ found\ a\ large\ elephant\| \tag{22}$$

$$\Rightarrow (\exists^1 john::\texttt{human})(\exists e::\texttt{elephant})$$
$$\quad (Found(j::\texttt{human}, e::\texttt{entity}) \wedge Large(e::\texttt{physical}))$$

$$\Rightarrow (\exists^1 john::(\texttt{human} \bullet \texttt{human}))(\exists e::(\texttt{elephant} \bullet (\texttt{entity} \bullet \texttt{physical})))$$
$$\quad (Found(j,e) \wedge Large(e))$$

$$\Rightarrow (\exists^1 john::\texttt{human})(\exists e::\texttt{elephant})(Found(j,e)) \wedge Large(e))$$

In the final analysis, therefore, if '*john found a large elephant*' then there is a specific `human` (named *j*), and some `elephant` *e*, such that *e* is *Large* and *j* found *e*. However, consider now the interpretation in (23).

$$\|john\ painted\ a\ large\ elephant\| \tag{23}$$

$$\Rightarrow (\exists^1 john::\texttt{human})(\exists e::\texttt{elephant})$$
$$\quad (Painted(j::\texttt{human}, e::\texttt{painting}) \wedge Large(e::\texttt{physical}))$$

Note that what we now have is a quantified variable, *e*, that is supposed to be an object of type `elephant`, an object that is described by a property, where it is considered to be an object of type `physical`, and an object that is in a relation in which it is considered to be a `painting`. There are two pairs of type unifications that must now occur, namely (`elephant` • `painting`) and (`elephant` • `physical`), where, if we recall the type unification definition given in (2), the former would result in making the reference to *e* abstract and in the introduction of a new variable of type `painting`. This process would in the final analysis result in the following:

$$\|john\ painted\ a\ large\ elephant\| \tag{24}$$

$$\Rightarrow (\exists^1 john :: \texttt{human})(\exists e :: \texttt{elephant}^a)(\exists p :: \texttt{painting})$$
$$(Painted(j,p) \land PaintingOf(p,e) \land Large(e))$$

Note here that the interpretation correctly states that it is a (painted) elephant (that need not actually exist) that is *Large* and not the painting itself. Thus, '*john painted a large elephant*' is correctly interpreted as roughly meaning '*john made a painting of a large elephant*'.

In addition to handling the so-called intensional verbs, our approach seems to also appropriately handle other situations that, on the surface, seem to be addressing a different issue. For example, consider the following:

*john read the book and then he burned it.* (25)

In Asher and Pustejovsky (2005) it is argued that 'book' in this context must have what is called a dot type, which is a complex structure that in a sense carries within it the semantic types associated with various senses of 'book'. For instance, it is argued that 'book' in (25) carries the `informational content' sense (when it is being read) as well as the `physical object' sense (when it is being burned). Elaborate machinery is then introduced to 'pick out' the right sense in the right context, and all in a well-typed compositional logic.

But this approach presupposes that one can enumerate, a priori, all possible uses of the word `book' in ordinary language[6]. Moreover, this approach does not seem to provide a solution for the problem posed by example (24), since there does not seem to be an obvious reason why a complex dot type for 'elephant' should contain a representational sense, although it is an object that can be painted. To see how this problem is dealt with in our approach, consider the following:

$$Read(x :: \texttt{human}, y :: \texttt{content}) \tag{26}$$
$$Burn(x :: \texttt{human}, y :: \texttt{physical}) \tag{27}$$

That is, we are assuming here that it always makes sense to speak of a human that read some content, and of a human that burned some physical object. Consider now the following:

$$\|john\ read\ a\ book\| \tag{28}$$

$$\Rightarrow (\exists^1 j :: \texttt{human})(\exists b :: \texttt{book})(Read(j :: \texttt{human}, b :: \texttt{content}))$$
$$\Rightarrow (\exists^1 j :: (\texttt{human} \bullet \texttt{human}))(\exists b :: (\texttt{book} \bullet \texttt{content}))(Read(j,b))$$
$$\Rightarrow (\exists^1 j :: \texttt{human})(\exists b :: \texttt{book})(\exists c :: \texttt{content})(ContentOf(b,c) \land Read(j,b))$$

Thus, if '*john read a book*' then there is some specific human (named *j*), some object *b* of type book, such that that *j* read the content of *b*. On the other hand, consider now the following:

---

[6] Similar presuppositions are also made in a hybrid (connectionist/symbolic) 'sense modulation' approach described by Rais-Ghasem and Corriveau (1998).

$$\|john\ burned\ a\ book\| \tag{29}$$
$$\Rightarrow (\exists^1 j :: \texttt{human})(\exists b :: \texttt{book})(Burn(j :: \texttt{human}, b :: \texttt{physical}))$$
$$\Rightarrow (\exists^1 j :: (\texttt{human} \bullet \texttt{human}))(\exists b :: (\texttt{book} \bullet \texttt{physical}))(Burned(j,b))$$
$$\Rightarrow (\exists^1 j :: \texttt{human})(\exists b :: \texttt{book})(Burned(j,b))$$

That is, if '*john burned a book*' then there is some specific human (named *j*), some object *b* of type book, such that that *j* burned *b*. Note, therefore, that when the book is being burned we are simply referring to the book as the physical object that it is, while reading the book implies, implicitly, that we are referring to an additional (abstract) object, namely the content of the book. The important point we wish to make here is that there is one book object, an object that is (ultimately) a physical object, that one can read (its content), sell/trade/etc (as a commodity), etc., or burn (as is, i.e., as simply the physical object that it is!) Thus 'book' can be easily used in different ways in the same linguistic context, as illustrated by the following:

$$\|john\ read\ a\ book\ and\ then\ he\ burned\ it\| \tag{30}$$
$$\Rightarrow (\exists^1 j :: \texttt{human})(\exists b :: \texttt{book})(Read(j :: \texttt{human}, b :: \texttt{content})$$
$$\Rightarrow (\exists^1 j :: \texttt{human})(\exists b :: \texttt{book})(Burn(j :: \texttt{human}, b :: \texttt{physical}))$$
$$\Rightarrow (\exists^1 j :: (\texttt{human} \bullet \texttt{human}))(\exists b :: (\texttt{book} \bullet \texttt{physical}))(Burned(j,b))$$
$$\quad \wedge ContentOf(b,c) \wedge Burn(j :: \texttt{human}, b :: \texttt{physical}))$$

Like the example of 'painting a large elephant' discussed in (24) above, where the painting of an elephant implied its existence in some painting and it being large as some physical object (that need not actually exist), in (30) we also have a reference to a book as a physical object (that has been burned), and to a book that has content (that has been read). Similar to the process depicted in figure 1 above, the type unifications in (30) should now result in the following:

$$\|john\ read\ a\ book\ and\ then\ he\ burned\ it\| \tag{31}$$
$$\Rightarrow (\exists^1 j :: \texttt{human})(\exists b :: \texttt{book})(\exists c :: \texttt{content})$$
$$\quad (ContentOf(c,b) \wedge Read(j,c) \wedge Burn(j,b))$$

That is, there is some unique object of type human (named *j*), some book *b*, some content *c*, such that *c* is the content of *b*, and such that *j* read *c* and burned *b*. As pointed out in a previous section, it should also be noted here that these type unifications are often retracted in the presence of additional information. For example, in '*John borrowed Das Kapital from the library. He did not agree with it*' the resolution of `it' would eventually result in the introduction of an abstract content object (which one might not agree with), as one does not agree (or disagree) with a physical object, an object that can indeed be borrowed[7].

---

[7] Interestingly, the resolution of 'it', in addition to introducing a content object, should result in *Das Kapital*, since one cannot agree or disagree with a library, but with the content of the library's books.

## 6 Towards *Discovering* the Structure of Commonsense Knowledge

Throughout this paper we have tried to demonstrate that a number of challenges in the semantics of natural language can be easily tackled if semantics is grounded in a strongly-typed ontology that reflects our commonsense view of the world and the way we talk about it in ordinary language. Our ultimate goal, however, is the systematic *discovery* of this ontological structure, and, as also argued in [2], [4] and [11], it is the systematic investigation of how ordinary language is used in everyday discourse that will help us discover (as opposed to invent) the ontological structure that seems to underlie all what we say in our everyday discourse. Recall, for example, our suggestion of how a nominal compound such as 'a book review' is interpreted:

$$\|a\ book\ review\| \Rightarrow \lambda P[(\exists x :: \texttt{book})(\exists y :: \texttt{review})(ReviewOf(y, x) \wedge P(y))] \tag{31}$$

Note that this analysis itself seems to shed some light on the nature of the ontological categories under consideration. For example, (31) seems to be an instance of a more generic template that can adequately represent the compositional meaning of a number of similar nominal compounds, as illustrated in (*a*) below.

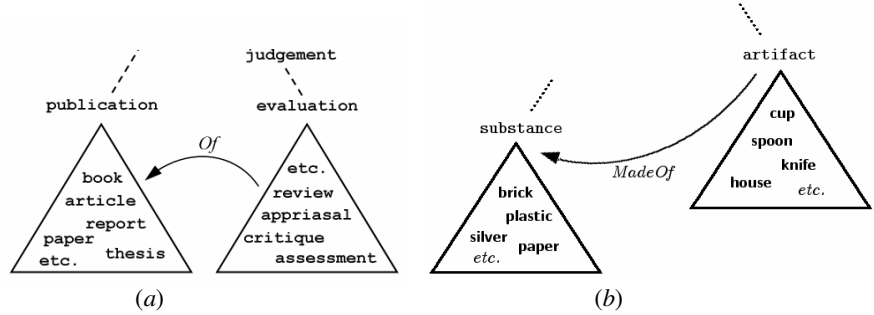

(*a*)                                   (*b*)

Similarly, as suggested in (*b*) above, it seems that the semantic analysis of a nominal compound such as 'brick house', for example, suggests that a number of nominal compounds can be semantically analyzed according to the following template:

$$\|a\ N_{\text{substance}}\ N_{\text{artifact}}\|$$
$$\Rightarrow \lambda P[(\exists x :: \texttt{artifact})(\exists y :: \texttt{substance})(MadeOf(x, y) \wedge P(x))]$$

The general strategy we are advocating can therefore be summarized as follows: (*i*) we can start our semantic analysis by assuming a set of ontological categories that are embedded in the appropriate properties and relations (based on our use of ordinary language); (*ii*) further semantic analysis of some non-trivial phenomena (such as nominal compounds, intensional verbs, metonymy, etc.) should help us put some structure on the ontological categories assumed in step (*i*); and (*iii*) this additional structure is then iteratively used to repeat the entire process until, presumably, the nature of the ontological structure that seems to be implicit in everything we say on ordinary language is well understood.

## 7  Concluding Remarks

The thesis presented in this paper is the following: assuming the existence of an ontological structure that reflects our commonsense view of the world and the way we talk about in ordinary language would make the semantics of various natural language phenomena nearly trivial. Although we could not, for lack of space, fully demonstrate the utility of our approach, recent results we have obtained suggest an adequate treatment of a number of phenomena, such as the semantics of nominal compounds, lexical ambiguity, and the resolution of quantifier scope ambiguities, to name a few. More importantly, it seems that a gradual imbedding of commonsense metaphysics in our semantic formalism is itself what will help us discover the nature of the ontological structure that seems to underlie all that we say in our every discourse.